\documentclass[runningheads]{llncs}
\usepackage[T1]{fontenc}
\usepackage{float} 
\usepackage{multirow}
\usepackage{bm}
\usepackage{amsmath}
\usepackage{mathtools}
\usepackage{amssymb}
\usepackage{booktabs}
\usepackage[table,xcdraw]{xcolor}
\usepackage{graphicx}
\usepackage{color}
\usepackage{wrapfig}
\definecolor{myblue}{rgb}{0.21,0.49,0.74}
\usepackage[pagebackref,breaklinks,colorlinks,allcolors=myblue]{hyperref}

\newcommand{\qsd}{\textcolor[rgb]{0,0,0}}

\begin{document}
%
\title{Skeleton-Snippet Contrastive Learning with Multiscale Feature Fusion for Action Localization}

%

\author{Qiushuo Cheng\inst{1}\and
Jingjing Liu\inst{1}\and
Catherine Morgan\inst{2} \and \\
Alan Whone\inst{2}\and
Majid Mirmehdi\inst{1}}

\authorrunning{Q. Cheng et al.}
%
\institute{School of Computer Science, University of Bristol\and North Bristol NHS Trust}
\maketitle              
\begin{abstract}
The self-supervised pretraining paradigm has achieved great success in learning 3D action representations for skeleton-based action recognition using contrastive learning. However, learning effective representations for skeleton-based temporal action localization  remains challenging and underexplored. Unlike video-level {action} recognition, detecting action boundaries requires temporally sensitive features that capture subtle differences between adjacent frames where labels change. 
To this end, we formulate a snippet discrimination pretext task for self-supervised pretraining, which densely projects skeleton sequences into non-overlapping segments and promotes features that distinguish them across videos via contrastive learning. Additionally, we build on strong backbones of skeleton-based action recognition models by fusing intermediate features with a U-shaped module to enhance feature resolution for frame-level localization.  
Our approach consistently improves existing skeleton-based contrastive learning methods for action localization on BABEL across diverse subsets and evaluation protocols.
{We also achieve state-of-the-art transfer learning performance on PKUMMD with pretraining on NTU RGB+D and BABEL. The code is available at https://github.com/cqs0925/SnipCLR.} 


\end{abstract}

\section{Introduction}
\label{sec:intro}
Skeleton sequences provide a lightweight approach to model human actions in various real-world applications, such as healthcare~\cite{review23healthapplication,majid2023remap,majid25yourturn} 
and surveillance~\cite{uav21}. For skeleton-based action recognition, traditional fully supervised learning requires large-scale training data with \textit{per-video} action labels~\cite{ntu16}.  
However, skeleton-based action localization requires \textit{per-frame} action labels, as it involves not only classifying actions but also detecting their temporal boundaries within the video~\cite{hanyuan,SWTAL23,SMQ25}. 
There are approaches using image labelling software to reduce the time burden and increase the accuracy of human-rater created manual labels \cite{Sager03072021}, but ultimately obtaining such manual labels is expensive, time-consuming and presents scalability challenges without significant resources. 

{To reduce reliance on manual labels, 3D action representation learning~\cite{skeletonCLR21,skeletonmae23} provides an effective solution through a two-stage paradigm: self-supervised pretraining followed by finetuning. 
Recently, this paradigm has achieved remarkable performance in skeleton-based \textit{action recognition} with the development of skeleton-based contrastive learning (CL) methods~\cite{skeletonCLR21,CDCLR22,aimclr22,RVTCLR,maskclr24,logoCLR24}. 
Such works apply global projections, such as global average pooling, after the backbone encoder to produce sequence-level representations for instance discrimination~\cite{instdisc18} in self-supervised pretraining. These sequence-level representations are then used as input to a classification head for video-level recognition during fine-tuning. Globally averaging over time causes the learned representation to be insensitive to frame-wise changes, while such temporal sensitivity~\cite{tsp21,BSP21} is critical for \textit{action localization}.}

\begin{figure}[t]
  \centering
   \includegraphics[width=.90\linewidth]{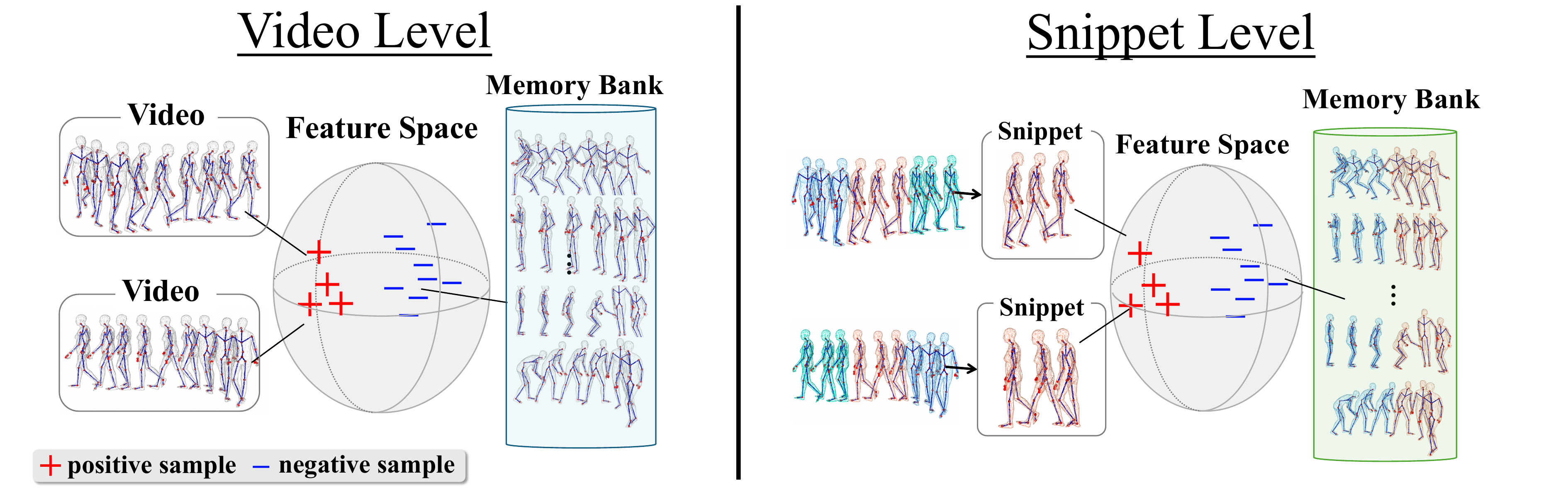}
   \caption{{\bf The basic concept of our contrastive approach.} {(Left)} Existing methods treat an entire skeleton sequence as a single sample in contrastive learning. {(Right)} Our approach delves into a more fine-grained level by considering snippets.}
   \label{fig:illustration}
   \vspace{-15pt}
\end{figure}


To address this, we draw inspiration from CL frameworks for dense prediction in images~\cite{densecl21,pixelpro21}, and extend them to 3D skeleton data by formulating a snippet discrimination pretext task to promote temporally distinct features.  
{In existing CL-based methods, snippets have been {obtained by, (i) using class labels to impose semantic relations between target actions or background~\cite{cola21,SSCL-T23,spcc_net24}, and (ii) sampling without labels and contrasting} them through heuristic relations~\cite{hitrs22,IICv222,RVTCLR} or self-supervised proxy tasks~\cite{hitrs22,tcgl22}  for representation learning. {Although also self-supervised, our method is unique in its dense use of snippets because it distinguishes among a large memory bank of finely divided snippets across the dataset (see Fig. \ref{fig:illustration}), which in turn makes the model more temporally aware of different action phases.} To construct snippet-level positive pairs, we determine dense correspondences between two augmented views of the same video using a similarity-based matching strategy~\cite{densecl21}.
To the best of our knowledge, we are the first to learn from untrimmed, unlabeled skeleton sequences for the dense downstream task of action localization. 
}

The field of skeleton-based {action recognition} has produced a rich set of powerful models with strong backbones and naturally multiscale intermediate features, including the successful spatial-temporal graph convolutional network (ST-GCN) family~\cite{2sagcn19,proto25}, 
as well as hierarchical transformer models~\cite{unik21,skateformer24,Hypergraphtransformer25}. 
We extend the capability of such backbones via a plug-and-play U-shaped module, {which upsamples high-level features progressively to recover temporal resolution, and leverages skip connections from intermediate features to preserve semantic information and fine-grained details.}


{In summary, the contributions of this work are: (i) we formulate snippet discrimination as a self-supervised proxy task to learn temporally-sensitive representations on untrimmed skeleton sequences, (ii) we introduce a plug-and-play U-shaped module to complement existing skeleton-based feature encoders for restoring feature resolution, (iii) we present extensive experiments and ablations to show that recent skeleton-based CL methods benefit from our approach in the downstream task of action localization.}

\section{Related Works}
\label{sec:background}
\noindent{\bf Self-supervised Learning with Skeletons --} 
\label{sec:ssl_review}
{Such methods can be broadly classified into two categories: (i)  generative methods which aim to reconstruct corrupted skeleton sequences~\cite{skeletonmae23,macdiff24} 
to learn spatial-temporal relations, and (ii) contrastive methods} which use transformations to build different `views' of the same skeleton, while pulling their encoded features close in latent space to facilitate the learning of high-level semantics. These methods mainly differ in their transformation strategies, including deriving other modalities (joint, bone, and motion)~\cite{skeletonCLR21}, extremely strong augmentations 
\cite{aimclr22}, relative playback speed sampling~\cite{RVTCLR}, gradual transformations~\cite{hiclr23}, adaptive masking~\cite{maskclr24}, local cropping~\cite{logoCLR24}, hierarchical downsampling~\cite{hico23} and  heterogeneity~\cite{heterCLR25}. LAC~\cite{LAC23} focuses on a related task of skeleton-based action segmentation by synthesizing skeleton data sequences and  encouraging similarities between frame-wise features of video-level positive pairs. In contrast, our method differs 
in that positive and negative samples are defined using snippets directly instead of videos. 

\noindent{\bf {Snippet-level Contrast for Representation Learning --}}
\label{sec:snipt_review}
Snippet-level contrastive learning typically extracts clips through cropping~\cite{logoCLR24} or sampling~\cite{iicv1} and defines contrastive pairs using proxy pretext tasks.
Some pretext tasks aim to preserve semantic consistency by defining positive pairs, such as local-to-global~\cite{maclearning22,logoCLR24} or snippet-to-snippet relations~\cite{iicv1} 
within the same video. Others focus on modeling temporal dependency by reordering the shuffled snippets~\cite{tcgl22} or predicting the next clip~\cite{hitrs22}, where positive pairs reflect the correct temporal order. Some RGB-based methods~\cite{iicv1,mamico22} 
focus on discriminating snippets across video instances, while snippets within the same video are usually treated as positives for they share the same visual context cues such as background. 
{Unlike RGB clips, intra-video skeleton snippets do not share background information and provide only an abstraction of semantically distinct sub-actions, which could be contrasted as negatives.} {Although a few RGB-based {action recognition}  methods~\cite{TCLR22,IICv222}} 
{also build intra-video negatives by using additional strong augmentations to neutralize shared backgrounds, their purpose is to highlight 
details for video-level recognition, especially between confusing action classes.}
{The skeleton-based CdCLR~\cite{CDCLR22} samples additional clips, but solely to augment the training data where each clip ranges from a short snippet to a full video.}

{Different from prior snippet-level contrastive methods, 
we target frame-level skeleton action localization rather than action recognition, where temporal information is important, but overlooked in current methods. Further, our snippets are uniformly divided and contrasted both within and across videos to promote temporally discriminative representations.} 

\noindent{\bf Multiscale Feature Learning with Skeletons --}
\label{sec:unet_review}
Multiscale architectures, e.g. U-Net~\cite{U-net} and feature pyramid networks~\cite{fpn}, are widely used and effective for diverse RGB-based dense prediction tasks, such as segmentation and localization. Following their success, these designs have been extended to skeleton data, e.g. combining U-Net with ST-GCN~\cite{uGCN19,weaklyUGCN} or 
transformers~\cite{uTRANSFORMER22,uTRANSFORMER24} to capture short- and long-term dependencies in 2D keypoint sequences for pose estimation. To model {actions at various temporal scales} for skeleton-based action segmentation, \cite{ghosh2020stacked} applies a decoder to ST-GCN to form an hourglass architecture, and then stacks multiple such modules to combine features at different scales. Some methods adopt a temporal pyramidal design, e.g. \cite{MST-GCN21,jang2024multi} 
encode multiscale skeleton features in parallel with early fusion, whereas~\cite{chen2025multiscale} performs late fusion on the final predictions. Though our {U-shaped} module 
is conceptually similar to these approaches, it differs in that it serves as a simple plug-in to existing skeleton-based action recognition backbones to enable dense action localization.

\section{Proposed Approach} \label{sec:methods}

{To support action localization in skeleton sequences, we improve upon existing MoCo-based video-level skeleton CL frameworks~\cite{skeletonCLR21,aimclr22,RVTCLR}. To preserve temporal sensitivity, we densely project each skeleton sequence into snippet-level embeddings to form positive and negative pairs, enabling an additional contrastive objective over snippets within and across videos. The formulation of positive and negative samples directly shapes the semantic information the model learns, thereby influencing downstream task performance.}

Given an input skeleton sequence $x$, different augmentations are applied to obtain the anchor–positive pair $(\bar{x}, \hat{x})$ as two semantically consistent views of $x$. Negative samples are obtained by sampling or mining~\cite{skeletonCLR21} from the remaining data.  
A first-in-first-out queue is used as a memory bank $\mathbf{M} = \{ q_n \}_{n=1}^M$ to store negative embeddings $q_n$ from previous batches. This design decouples the memory-bank size from the batch size, allowing us to set  $M=32,768$, which is sufficiently large to provide stable performance~\cite{skeletonCLR21,aimclr22}. 

\begin{figure*}[t]
  \centering
   \includegraphics[width=0.9\linewidth]{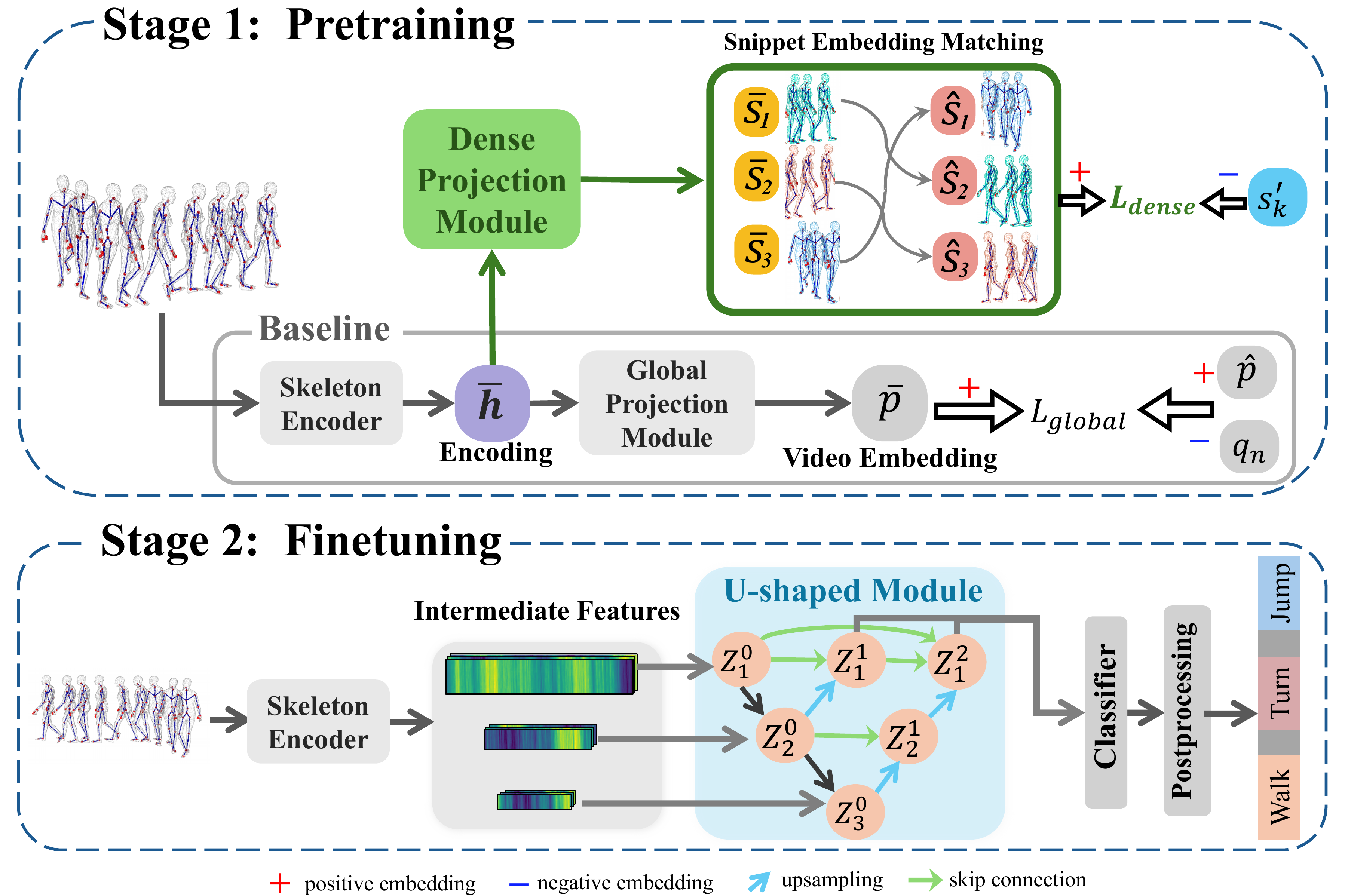}
   \caption{{\bf Overall pipeline of pretraining and finetuning in  proposed designs.} 
   {In Stage 1, we build upon the the existing video-level skeleton-based {CL} {baselines} with a Dense Projection Module to obtain snippet-level skeleton embeddings for fine-grained contrastive learning, aligning matched snippets as positives while separating negatives.
   In Stage 2, a U-shaped module is plugged into the pretrained skeleton encoder to restore temporal resolution while fusing intermediate features through skip connections.}}
     
   \label{fig:main}
   \vspace{-15pt}
\end{figure*}

{Our approach follows a two-stage paradigm (see Fig. \ref{fig:main}) commonly adopted by existing methods~\cite{skeletonCLR21,aimclr22,RVTCLR,maskclr24,heterCLR25}.} 
{In Stage 1, we integrate our dense snippet-level contrastive objective in self-supervised contrastive pretraining. In Stage 2, a U-shaped module is plugged into the pretrained encoder to fuse multiscale features during the finetuning for skeleton-based action localization under various evaluation settings.}

\subsection{Stage 1: Dense Snippet Contrast Pretraining}
\label{sec:dense contrast}
{Following~\cite{skeletonCLR21,aimclr22,RVTCLR}, we use an encoder $\mathcal{E}$ to obtain the skeleton encoding $\bar{h} = \mathcal{E}(\bar{x})$ from $\bar{x}$. We then employ a global projection module $\mathcal{G}$ and a dense projection module $\mathcal{D}$ separately, to  produce video-level and snippet-level embeddings.}

\noindent{\bf Global Projection Module --} {The global projection module $\mathcal{G}$ takes the skeleton encoding $\bar{h}$ as input, performs global pooling over time, and {uses a linear layer as a linear projection to adjust the channel dimension.}  {The operations in $\mathcal{G}$ can be} formally defined as}

\begin{equation}
\label{eq:global_projector}
\bar{h} \in \mathbb{R}^{T \times C_e} 
\xrightarrow{\text{global pooling}} 
\bar{F}_g \in \mathbb{R}^{C_e} 
\xrightarrow{\text{linear proj.}} 
\bar{p} \in \mathbb{R}^{C},
\end{equation}
where $\bar{F}_g$ denotes the globally aggregated representation, $\bar{p}$ is the video-level anchor embedding, $T$ and $C_e$ are the temporal length and channel dimension of {the skeleton encoding}, and $C$ are the channels of the video-level embedding. {The momentum-updated duplicates of the skeleton encoder and the global projection module are used similarly to obtain the positive embedding $\hat{p}$ from $\hat{x}$}. 
To ensure consistency across batches, {video-level negative embeddings} $q_n$ are obtained in the same manner. 
The InfoNCE loss~\cite{moco20} 
is then applied to contrast these embeddings as
\begin{equation}
\label{eq:infonce}
\mathcal{L}_{\text{global}} = -\log \frac{\exp(\bar{p} \cdot \hat{p} / \tau)}{\exp(\bar{p} \cdot \hat{p} / \tau) + \sum_{n=1}^{M} \exp(\bar{p} \cdot q_n / \tau)},
\end{equation}
where $\tau$ is the temperature hyperparameter.

\noindent{\bf Dense Projection Module --}
We extend the contrastive objective in Eq.~\ref{eq:infonce} to the snippet level by introducing a dense projection module $\mathcal{D}$. 
The concept of leveraging dense projection in contrastive learning originates from RGB-based studies~\cite{densecl21,pixelpro21}, and we adapt it to skeleton data. {The skeleton encoding $\bar{h}$ is divided into $N$ temporal snippets via adaptive local pooling, and then projected to a shared embedding dimension $C$ using a convolutional projection head built with two 1$\times$1 convolutions and a ReLU activation in between. The complete dense projection module $\mathcal{D}$ is formally defined as}
\begingroup
\begin{equation}
\label{eq:dense_projector}
\bar{h} \in \mathbb{R}^{T \times C_e} 
\xrightarrow{\text{local pooling}} 
\bar{F}_d \in \mathbb{R}^{N \times C_e} 
\xrightarrow{\text{conv. proj.}} 
\bar{S} \in \mathbb{R}^{N \times C},
\end{equation}
\endgroup
\noindent{where $\bar{F_d}$ are the {snippet-level representations} after pooling, and $\bar{S} \in \mathbb{R}^{N \times C}$ denotes the snippet-level embeddings in the anchor sample $\bar{x}$.} Similarly, the momentum-updated duplicate {of the dense projection module}  is used to obtain the {snippet embeddings $\hat{S}$ from positive sample} and {snippet embeddings $S'$ from negative samples.} {Having obtained these embeddings via the dense projection module, next we describe the snippet-level positive embedding matching between $\bar{S}$ and $\hat{S}$, and the construction of snippet-level negative embeddings $s^\prime_k$ from $S^\prime$. }


\noindent{\bf Constructing Snippet-level Positive and Negative Embeddings --}
Considering various strong spatial and temporal transformations are applied on the skeleton sequence, there is no guarantee that snippets at the transformed temporal locations across views would remain semantically relevant as the positive pair. 
\begin{wrapfigure}{r}{0.3\textwidth}
\vspace{-25pt}
\centering
   \includegraphics[width=0.3\textwidth]{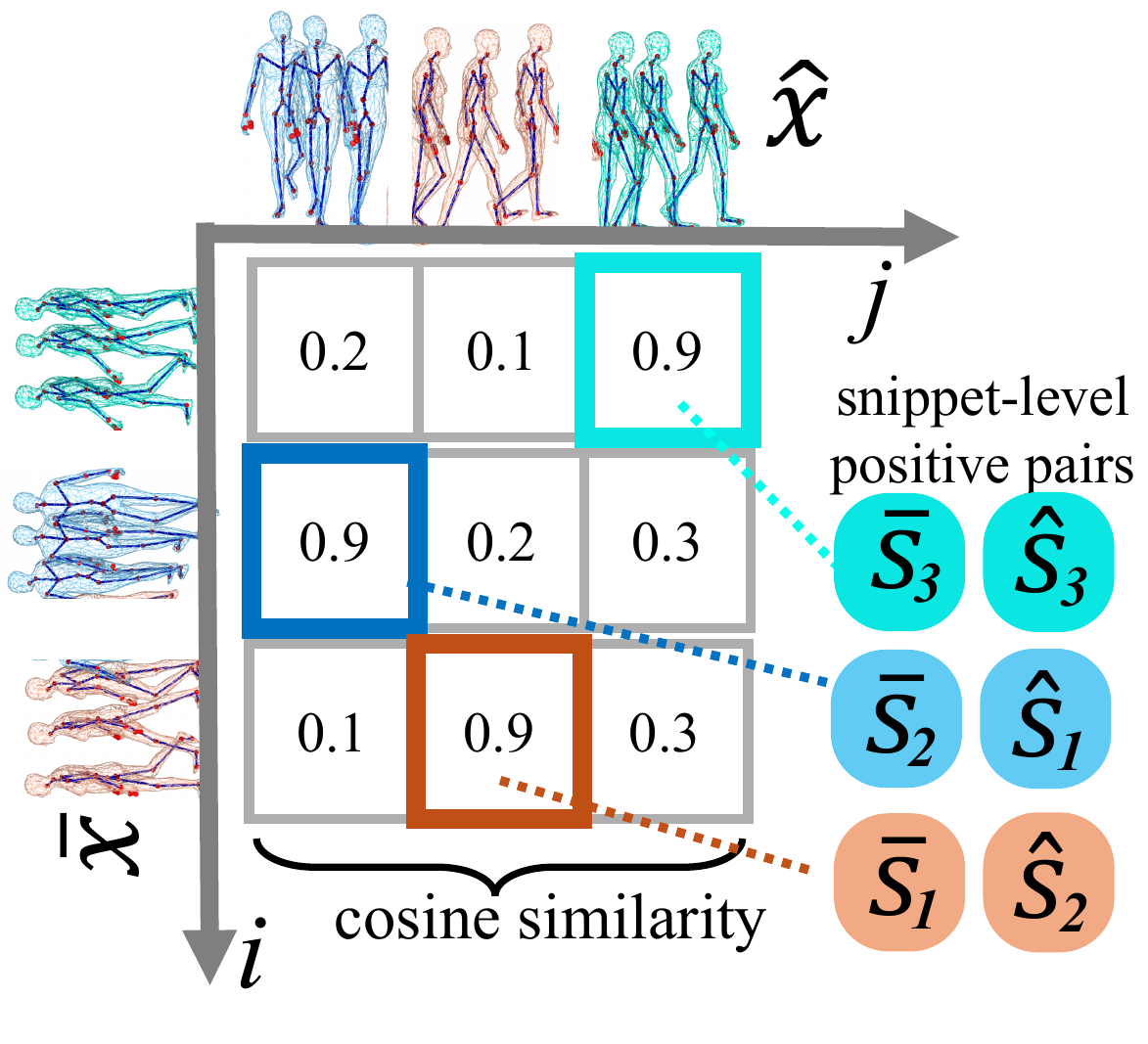}
   \caption{{Similarity-based matching.}}
   \label{fig:matching}
   \vspace{-25pt}
\end{wrapfigure} 
In our pretext task, each snippet is treated as a distinct instance in its own right to discriminate against other snippets across videos. 
We therefore adopt a similarity-based matching strategy~\cite{densecl21} to establish temporal correspondence between snippets in 
different views (see Fig.~\ref{fig:main}). 
For each {snippet embedding $\bar{s_i} \in \bar{S}$}, its feature is $\bar{f_i} \in \mathbb{R}^{C_e}$, corresponding to the $i^{th}$ row of the snippet-level representations $\bar{F_d} \in \mathbb{R}^{N \times C_e}$. To identify the matched {positive snippet embedding $\hat{s} \in \hat{S}$} of the other view $\hat{x}$, we select the snippet whose feature $\hat{f_j}$ is most similar to $\bar{f_i}$, such that 
\begin{equation}
j^\ast = \mathop{\arg\max}_{j}  \mathrm{sim}(\bar{{f}_i}, \hat{{f}_j}),
\label{eq:match}
\end{equation}
where $\mathrm{sim}$ computes the cosine similarity, and $\hat{s}_{j^\ast}$ is the matched positive snippet embedding for $\bar{s_i}$ {(see Fig. \ref{fig:matching}).} To construct {snippet-level negative embeddings $s^\prime_k \in \mathbb{R}^{C}$,} 
a natural idea is to sample snippets from other samples. In practice, we found that using global pooling on $S^\prime \in \mathbb{R}^{N \times C}$ yields slightly better results, so we adopt this strategy. 
Negative embeddings in the current batch are enqueued, while the earliest embeddings in $\mathbf{M^\prime}=\{ s^\prime_k \}_{k=1}^M$ are dequeued. The dense temporal contrastive loss can now be calculated as
\begin{equation}
\label{eq:dense_loss}
\mathcal{L}_{\text{dense}} =\sum ^N_{i=0}-\log \frac{\exp(\bar{s_i} \cdot \hat{s}_{j^\ast} / \tau)}{\exp(\bar{s}_i \cdot \hat{s}_{j^\ast} / \tau) + \sum_{k=1}^{M} \exp(\bar{s}_i \cdot s^\prime_k / \tau)}.
\end{equation}
Finally, the total loss is 
\begin{equation}
\label{eq:total_loss}
\mathcal{L} = \mathcal{L}_{\text{global}} + \lambda \mathcal{L}_{\text{dense}},
\end{equation}
where $\lambda$ is the hyperparameter balancing these two objectives. {We ablate the value $\lambda$ in Section \ref{sec:Experiments} to study its influence on performance.}

\subsection{Stage 2: Multiscale Feature Fusion Finetuning}
\label{sec: unet}
{After pretraining, we retain only the skeleton encoder as the backbone and finetune it for skeleton-based action localization, leveraging multiscale feature fusion.
To achieve this, }we utilise a U-shape structure~\cite{U-net} with nested skip connections~\cite{unet++} {to fuse features at different scales} and progressively upsample the final output to the original temporal resolution.
{Let {$ a \in [1, L) $} denote the index of the $a^{th}$ downsampling level in the encoder, and $b$ the index of 
the downsampling or upsampling operation at level $a$. We use $Z_a^b$ (see Fig. \ref{fig:main}) to denote each feature in the proposed U-shaped structure.} When $b = 0$, $Z_a^0$ represents the intermediate feature produced by the encoder during downsampling, such that
\begin{equation}
Z_{a+1} ^0 = \mathcal{R}(Z_a^0),
\end{equation}
where $\mathcal{R}$ is the downsampling module in the encoder. When $b > 0$, $Z_a^b$ represents the upsampled feature map along the skip connections pathway, computed as
\begin{equation}
Z_a^b = 
\mathcal{H}\left(
    \mathrm{concat}\left(
        \left[ Z_a^0, Z_a^1, \dots, Z_a^{b-1},\ \mathcal{U}(Z_{a+1}^{b-1}) \right]
    \right)
\right).
\end{equation}
Here, $\mathrm{concat}$ denotes the skip connection, which concatenates features along the channel dimension, $\mathcal{H} $ represents a simple Conv1D projection for channel adjustment, and $\mathcal{U}$ is the upsampling operation that restores temporal resolution. While in semantic segmentation  the number of channels is typically reduced to match the number of output classes, we maintain a hidden dimension $d=512$ for channel size to enhance the ability to capture rich features as frame-level action representations \cite{bsn++21}. {Finally, we attach a classifier (see Section~\ref{sec:Experiments}) to produce frame-wise logits and apply the postprocessing method from~\cite{SWTAL23}, which uses multiple thresholds to generate predicted action instances followed by non-maximum suppression to remove overlapping segments.}

\section{Experiments}
\label{sec:Experiments}

\noindent{\bf Datasets --} 
BABEL~\cite{babel21} is a large-scale 3D human motion dataset of \textit{untrimmed} sequences containing multiple actions.
It comprises {43.5} hours of mocap data from AMASS~\cite{mahmood2019amass} with over 63k frame-level annotation labels. 
We perform pretraining on the official BABEL training split, unless otherwise specified. Due to the low-data regime, we follow the protocol in~\cite{SWTAL23,SMQ25} to evaluate the performance of self-, weakly-, and semi-supervised by finetuning on three subsets of the BABEL validation split.
{Specifically, Subset-1 includes \textit{Walk}, \textit{Stand}, \textit{Turn}, and \textit{Jump}, Subset-2 includes \textit{Sit}, \textit{Run}, \textit{Stand-up}, and \textit{Kick}, and Subset-3 includes \textit{Jog}, \textit{Wave}, \textit{Dance}, and \textit{Gesture}.}
{To study the transfer learning performance from different types of pretraining datasets {(see Tab. \ref{tab:data_statistics})}, we also pretrain on the NTU RGB-D dataset (NTU-60)~\cite{ntu16}. NTU-60 contains  56,578 \textit{curated,} trimmed videos with 60 actions,  performed by 40 actors and captured by three Microsoft Kinect v2 cameras.  
Following the transfer learning protocol of~\cite{hitrs22,USDRL25}, we use the cross subject training split of NTU-60, and finetune on PKU Multi-Modality Dataset (PKUMMD)~\cite{PKU}, which consists of 1076 long video sequences, with 51 action categories and {21545} action instances.}

\begin{wraptable}{r}{0.54\textwidth}
\vspace{-5pt}
\scriptsize
\caption{{Pretraining data statistics with {clip duration} timings in seconds.}}
\centering
\begin{tabular}{l|rr|rrrrrr}
\toprule
    \textbf{Dataset} & \textbf{\#Videos} &\textbf{\#Frames} &\textbf{$t_\text{mean}$} & \textbf{$t_\text{std}$} & \textbf{$t_\text{min}$} & \textbf{$t_\text{max}$} \\
    \midrule
    BABEL   & 6613 & 2.4M & 11.9  & 16.2&0.1  &  263.6  \\
    NTU-60 & 40091 & 4.0M & 3.3 & 0.1 & 0.1& 3.3 \\
\bottomrule
    \end{tabular}
    \label{tab:data_statistics} 
    \vspace{-15pt}
\end{wraptable}

\noindent{\bf Implementation Details and Evaluation --} We verify our approach by integrating with three  skeleton-based CL methods as baselines, i.e. AimCLR~\cite{aimclr22}, HiCLR~\cite{hiclr23}, and RVTCLR~\cite{RVTCLR}.
We also report the transfer learning experiments results from NTU-RGB-D~\cite{ntu16} and BABEL to PKUMMD~\cite{PKU} with comparison to other state-of-the-art representation learning methods~\cite{hiclr23,PCM23,USDRL25}. {Unless otherwise noted, we use 2s-AGCN \cite{2sagcn19} as the shared backbone for all prior methods~\cite{aimclr22,hiclr23,RVTCLR} and ours to ensure a fair comparison.}
{Although CdCLR~\cite{CDCLR22} is conceptually the most similar approach, it is not open-sourced and lacks implementational details, hence we cannot compare against it.} 
For data preprocessing on the BABEL dataset, we follow \cite{SWTAL23} to extract 25 keypoints from SMPL parameters~\cite{babel21}, downsample the sequences to 30 fps, center the skeleton at the root joint, and align the spine and shoulders to the z- and  x-axes.
Memory-bank size $M=32768$ and temperature $\tau=0.007$ are set as in baseline methods.
{We set the number of snippet $N=19$, and ablated it to study its effect.}
The model is optimized using SGD with momentum (0.9) and weight decay (0.0001). 
For downstream evaluation, we report the mean Average Precision (mAP) under temporal Intersection over Union (tIoU) thresholds. We adopt three common evaluation settings, detailed below. 

\noindent{\bf KNN and Linear Evaluation on BABEL --}
We conduct two experiments on the frozen skeleton encodings: (i) applying a K-Nearest Neighbor (KNN) classifier, and (ii) finetuning a linear classifier. 
{These commonly adopted evaluation settings  reflect the quality of the pretrained feature directly, given that the backbone encoder is kept frozen during finetuning~\cite{moco20,skeletonCLR21,hiclr23}.} Under KNN evaluation, our method consistently improves average mAP across tIoU thresholds and all subsets for AimCLR~\cite{aimclr22}, HiCLR~\cite{hiclr23}, and RVTCLR~\cite{RVTCLR}. For example, as shown in Tab. \ref{tab:knn}, AimCLR increases from 37.1\% 
to 42.6\% ($\uparrow$5.5\%) on Subset-2, as does  HiCLR from 41.0\% 
to 43.4\% ($\uparrow$2.4\%); RVTCLR rises from 38.2\% 
to 43.4\% ($\uparrow$5.2\%) on Subset-1, and from 22.7\% 
up to 31.2\% ($\uparrow$8.5\%) on Subset-3. The only exception is HiCLR on Subset-1, which decreases slightly from 43.7\% 
to 42.7\% ($\downarrow$1.1\%).  

A similar trend holds under linear evaluation (Tab. \ref{tab:linear}), where our method reliably {boosts} mAP across subsets and thresholds. HiCLR and RVTCLR achieve substantial gains in mAP across all three subsets. For instance, HiCLR on Subset-2 {rises} from 41.1\% to 47.4\% ($\uparrow$6.3\%), and RVTCLR from 40.9\% to 45.3\% ($\uparrow$4.4\%) on Subset-1. For AimCLR, the average mAP {increases} from 43.3\% to 46.6\% ($\uparrow$3.3\%) on Subset-2, and from 38.6\% to 41.0\% ($\uparrow$2.4\%) on Subset-3, with only a minor drop from 42.6\% to 42.3\% ($\downarrow$0.3\%) on Subset-1. Overall, these results indicate that our pretraining produces higher-quality features for action localization, primarily due to the introduction of a more fine-grained contrastive objective\setcounter{footnote}{0}\footnote{\scriptsize  Additionally, we provide extensive KNN and linear evaluation results using different backbone architectures, including ST-GCN~\cite{stgcn18}, 2s-AGCN~\cite{2sagcn19} and UNIK~\cite{unik21}, in the Suppl. Material.}.

\begin{table*}[t]
\caption{{\bf KNN evaluation with different skeleton-based CL methods.} mAP averaged over actions at tIoU thresholds 0.1 to 0.5 on three subsets of BABEL.}
\centering
\renewcommand{\arraystretch}{1.3}
\setlength{\tabcolsep}{2pt}
\resizebox{\linewidth}{!}{
\begin{tabular}{c|ccccc>{\columncolor{gray!10}}c|ccccc>{\columncolor{gray!10}}c|ccccc>{\columncolor{gray!10}}c}
\toprule
\multirow{2}{*}{\textbf{Methods}}   
& \multicolumn{6}{c|}{\textbf{Subset-1}} & \multicolumn{6}{c|}{\textbf{Subset-2}} & \multicolumn{6}{c}{\textbf{Subset-3}}\\
& 0.1& 0.2& 0.3& 0.4& 0.5 & \textbf{Avg.} & 0.1& 0.2& 0.3& 0.4& 0.5& \textbf{Avg.} & 0.1& 0.2& 0.3& 0.4& 0.5 &\textbf{Avg.}\\
\midrule
AimCLR~\cite{aimclr22} 
& \textbf{51.7} & 46.6 & \textbf{42.3} & 35.1 & 28.5 & 40.8 
& 48.8 & 43.6 & 38.0 & 31.6 & 23.4 & 37.1 
& 34.2 & 29.5 & 25.4 & 22.4 & 18.3 & 25.9 \\

+Ours 
& 50.3 & \textbf{46.5} & 42.0 & \textbf{36.2} & \textbf{29.6} & \textbf{40.9} 
& \textbf{54.6} & \textbf{50.0} & \textbf{43.7} & \textbf{35.9} & \textbf{28.7} & \textbf{42.6} 
& \textbf{36.5} & \textbf{32.3} & \textbf{27.4} & \textbf{26.4} & \textbf{20.0} & \textbf{28.5} \\

\midrule
HiCLR~\cite{hiclr23} 
& \textbf{53.7} & \textbf{49.5} & \textbf{44.9} & \textbf{38.2} & 32.4 & \textbf{43.7} 
& 55.0 & 49.4 & 42.0 & 33.0 & 25.4 & 41.0 
& 31.5 & 27.2 & 21.4 & 17.4 & 13.7 & 22.2 \\

+Ours 
& 51.7 & 47.2 & 43.4 & 38.0 & \textbf{33.1} & 42.7 
& \textbf{56.7} & \textbf{51.2} & \textbf{43.8} & \textbf{35.5} & \textbf{29.6} & \textbf{43.4} 
& \textbf{31.9} & \textbf{27.7} & \textbf{24.8} & \textbf{21.3} & \textbf{19.2} & \textbf{25.0} \\

\midrule
RVTCLR~\cite{RVTCLR}  
& 48.2 & 43.4 & 39.6 & 33.1 & 26.8 & 38.2 
& 52.3 & 44.2 & 35.7 & 27.4 & 21.8 & 36.3 
& 29.2 & 24.4 & 21.6 & 20.2 & 18.0 & 22.7 \\

+Ours 
& \textbf{52.6} & \textbf{48.1} & \textbf{44.5} & \textbf{39.2} & \textbf{32.2} & \textbf{43.4} 
& \textbf{59.6} & \textbf{55.0} & \textbf{46.2} & \textbf{40.3} & \textbf{30.1} & \textbf{46.2} 
& \textbf{42.0} & \textbf{34.3} & \textbf{29.8} & \textbf{25.6} & \textbf{24.2} & \textbf{31.2} \\

\bottomrule
\end{tabular}}
\vspace{-5pt}
\label{tab:knn}
\end{table*}

\begin{table*}[t]
\caption{{\bf Linear evaluation with different skeleton-based CL methods.} mAP averaged over actions at tIoU 0.1 to 0.5  on three subsets of BABEL.}
\centering
\renewcommand{\arraystretch}{1.3}
\setlength{\tabcolsep}{2pt}
\centering
\resizebox{\linewidth}{!}{
\begin{tabular}{c|ccccc>{\columncolor{gray!10}}c|ccccc>{\columncolor{gray!10}}c|ccccc>{\columncolor{gray!10}}c}
\toprule
\multirow{2}{*}{\textbf{Methods}}     
& \multicolumn{6}{c|}{\textbf{Subset-1}} & \multicolumn{6}{c|}{\textbf{Subset-2}} & \multicolumn{6}{c}{\textbf{Subset-3}}\\
& 0.1& 0.2& 0.3& 0.4& 0.5 & \textbf{Avg.}& 0.1& 0.2& 0.3& 0.4& 0.5& \textbf{Avg.} & 0.1& 0.2& 0.3& 0.4& 0.5 &\textbf{Avg.}\\
\midrule
AimCLR~\cite{aimclr22} 
& \textbf{58.7} & \textbf{51.4} & \textbf{42.6} & 34.2 & 26.1 & \textbf{42.6} 
& 62.1 & 53.2 & 41.9 & 33.8 & 25.2 & 43.3 
& 46.9 & \textbf{45.0} & 38.2 & 32.7 & 30.4 & 38.6 \\

+Ours 
& 55.7 & 51.3 & 42.4 & \textbf{35.1} & \textbf{26.9} & 42.3 
& \textbf{64.2} & \textbf{55.4} & \textbf{46.4} & \textbf{38.6} & \textbf{28.3} & \textbf{46.6} 
& \textbf{48.5} & 44.0 & \textbf{42.3} & \textbf{37.5} & \textbf{32.9} & \textbf{41.0} \\

\midrule
HiCLR~\cite{hiclr23} 
& \textbf{57.7} & 51.5 & 42.0 & 34.2 & 25.1 & 43.3 
& 61.1 & 50.9 & 40.7 & 30.4 & 22.2 & 41.1 
& \textbf{46.5} & 42.8 & 34.9 & 31.0 & 28.3 & 36.7 \\

+Ours 
& 56.9 & \textbf{52.1} & \textbf{44.6} & \textbf{36.1} & \textbf{28.7} & \textbf{43.7} 
& \textbf{63.7} & \textbf{57.1} & \textbf{48.5} & \textbf{38.5} & \textbf{29.2} & \textbf{47.4} 
& 45.8 & 42.7 & \textbf{37.9} & \textbf{32.3} & \textbf{31.1} & \textbf{38.0} \\

\midrule
RVTCLR~\cite{RVTCLR} 
& 57.0 & 50.1 & 41.9 & 31.8 & 23.7 & 40.9 
& 57.7 & 48.5 & 38.2 & 27.9 & 20.7 & 38.8 
& 43.3 & 40.4 & 34.2 & 29.8 & 26.5 & 34.8 \\

+Ours 
& \textbf{60.5} & \textbf{55.0} & \textbf{45.8} & \textbf{36.6} & \textbf{28.8} & \textbf{45.3} 
& \textbf{65.2} & \textbf{57.8} & \textbf{49.0} & \textbf{40.4} & \textbf{32.9} & \textbf{49.1} 
& \textbf{47.9} & \textbf{46.8} & \textbf{41.7} & \textbf{31.7} & \textbf{28.2} & \textbf{39.3} \\

\bottomrule
\end{tabular}
}

\label{tab:linear}
\vspace{-10pt}
\end{table*}


\begin{table*}[ht]
\centering

\begin{minipage}{0.5\textwidth} 
\centering
\caption{\textbf{Weakly- and semi-supervised evaluation on BABEL.} {mAP averaged over tIoU thresholds 0.1 to 0.5 over 3 runs.}}
\resizebox{\linewidth}{!}{
\begin{tabular}{c|c|cccc}
\toprule
\multirow{2}{*}{\textbf{Pretraining}} & \multirow{2}{*}{\textbf{Setting}} & \multicolumn{4}{c}{\textbf{mAP@tIoU (\%)}}  \\ 
& & Subset-1 & Subset-2 & Subset-3 &   \cellcolor{gray!10}\textbf{Avg.}\\
\midrule

Random Init.  & Weak & 34.2 & 40.2 & 27.5 & \cellcolor{gray!10}{34.0} \\
Random Init.  & Full & 47.3 & 50.8 & 39.2 & \cellcolor{gray!10}{45.8} \\
\midrule

RVTCLR~\cite{RVTCLR} & Weak & 44.0 & 42.5 & \textbf{38.7} & \cellcolor{gray!10}{41.7} \\ 
\ + Ours & Weak & \textbf{46.8} & \textbf{44.2} & 36.8 & \cellcolor{gray!10}{\textbf{42.6}} \\
\midrule

RVTCLR~\cite{RVTCLR} & 10\% & 49.8 & \textbf{45.9} & \textbf{38.7} & \cellcolor{gray!10}{44.8} \\
\ + Ours & 10\% & \textbf{51.0} & 45.4 & 38.4 & \cellcolor{gray!10}{\textbf{44.9}} \\ 
\midrule

RVTCLR~\cite{RVTCLR} & 20\% & 51.9 & 50.9 & 40.3 & \cellcolor{gray!10}{47.7} \\
\ + Ours & 20\% & \textbf{52.1} & \textbf{51.5} & \textbf{42.5} & \cellcolor{gray!10}{\textbf{48.7}} \\
\midrule

RVTCLR~\cite{RVTCLR} & 30\% & 53.0 & 56.0 & 42.3 & \cellcolor{gray!10}{50.4} \\
\ + Ours & 30\% & \textbf{53.5} & \textbf{56.2} & \textbf{44.2} & \cellcolor{gray!10}{\textbf{51.3}} \\
\midrule

RVTCLR~\cite{RVTCLR} & 40\% & 52.4 & 56.2 & 42.4 & \cellcolor{gray!10}{50.3} \\
\ + Ours & 40\% & \textbf{53.9} & \textbf{57.7} & \textbf{45.2} & \cellcolor{gray!10}{\textbf{52.2}} \\

\bottomrule
\end{tabular}}
\label{tab:semi_weak}
\end{minipage}
\hfill
\begin{minipage}{0.45\textwidth} 
\centering
\caption{{\bf Transfer learning evaluation.} Using different pretraining data and methods on PKUMMD.}
\resizebox{\linewidth}{!}{
\begin{tabular}{c|c|c|c}
\toprule
\textbf{Model} & \textbf{Pretrain} & \textbf{Backbone}  & \textbf{mAP@0.5}\\
\midrule
Hi-TRS~\cite{hitrs22}         & \multirow{3}{*}{NTU-60}& FCV-TRS    & 66.6 \\
PCM$^3$~\cite{PCM23}          & & BiGRU      & 82.3 \\
USDRL~\cite{USDRL25}          &  & DSTE       & 77.4 \\
\midrule
RVTCLR~\cite{RVTCLR}          & \multirow{2}{*}{NTU-60} & \multirow{2}{*}{2s-AGCN}  & \textbf{84.1} \\
\ + Ours                      &  &     & 76.1 \\
\midrule
RVTCLR~\cite{RVTCLR}          & \multirow{2}{*}{BABEL}  & \multirow{2}{*}{ST-GCN}       & 78.0\\
\ + Ours                      &   &       & \textbf{80.6} \\
\midrule
RVTCLR~\cite{RVTCLR}          & \multirow{2}{*}{BABEL}   & \multirow{2}{*}{2s-AGCN}   & 84.6 \\
\ + Ours                      &   &  & \textbf{84.7} \\
\midrule
RVTCLR~\cite{RVTCLR}          & \multirow{2}{*}{BABEL}  & \multirow{2}{*}{UNIK}       & 85.3\\
\ + Ours                      &   &       & \textbf{86.0} \\
\midrule
RVTCLR~\cite{RVTCLR}          & \multirow{2}{*}{BABEL}  & \multirow{2}{*}{HyperGCN}       & 60.1\\
\ + Ours                      &  &       & \textbf{67.5} \\
\bottomrule
\end{tabular}}
\label{tab:transfer}
\end{minipage}
\vspace{-10pt}
\end{table*}

\noindent{\bf Weakly- and Semi-supervised Evaluation on BABEL --}
We pretrain the skeleton encoder {2s-AGCN~\cite{2sagcn19}} on all unlabeled data and append a linear classifier to the encoder. We finetune the entire model under two data regimes: (i) video-level weak labels~\cite{SWTAL23} for all samples, and (ii) 10\% or 20\% of randomly selected labeled samples. {A randomly initialized training-from-scratch model is also included as a baseline.} 
The weakly-supervised and semi-supervised results, presented in Tab. \ref{tab:semi_weak}, show that leveraging our method further reduces reliance on labeled data compared to the baseline pretraining method, RVTCLR~\cite{RVTCLR}. In weakly-supervsied evaluation, our method boosts the average mAP of baseline from 41.7\% to 42.6\% ($\uparrow$0.9\%). Notably, on Subset-1, using only weak labels achieves performance comparable to fully supervised training from scratch, with {average} mAP scores of 46.8\% and 47.3\%, respectively. 
For semi-supervised evaluation, finetuning with only 20\% of randomly selected samples outperforms training from scratch with all training samples, increasing the average mAP from 45.8\% to 48.7\% ($\uparrow$2.9\%). There is a minor performance drop in some cases. For example, on Subset-2 with 10\% of training data, the {average} mAP drops from 45.9\% to 45.4\% ($\downarrow$0.5\%).
{We attribute this to Subset-2 and Subset-3 being much smaller than Subset-1, making the 10\% split too small for stable finetuning. Aside from this variation, our method yields consistent improvements over the baseline  at 20\%, 30\%, and 40\% supervision.}

\noindent{\bf Transfer Learning Evaluation on PKUMMD --}
We follow~\cite{hitrs22,USDRL25} to evaluate transfer learning on the PKUMMD dataset. As shown in Tab.~\ref{tab:transfer}, our method improves on the baseline CL method RVTCLR~\cite{RVTCLR} across different backbones, increasing mAP from {78.0\% to 80.6\% ($\uparrow$2.6\%)} with ST-GCN~\cite{stgcn18},  {from 84.6\% to 84.7\% ($\uparrow$0.1\%)} with 2s-AGCN~\cite{2sagcn19}, {85.3\% to 86.0\% ($\uparrow$0.7\%)} with UNIK~\cite{unik21}, {and 60.1\% to 67.5\%($\uparrow$7.4\%) with HyperGCN~\cite{HyperGCN}.} {Although HyperGCN is more recent than ST-GCN, it performs worse on localization due to hyperedges aggregating global information which weakens intra-frame distinctiveness.}
Interestingly, when pretrained on NTU-60, our method leads to a noticeable performance drop, whereas pretraining on BABEL yields improvements. We attribute this to the nature of the datasets (see Tab.~\ref{tab:data_statistics}): NTU-60 contains short clips of less than 100 frames with a single curated action, therefore further fragmentation produces snippets too short to retain sufficient semantic meaning. 
In contrast, BABEL contains long and untrimmed sequences with multiple actions, which benefit from our dense contrastive projection. Our method is purpose-built for learning from  uncurated skeleton sequences without labels for localization, and such unconstrained data are more easily accessible while providing richer supervision signals.
Although BABEL is roughly half the size of NTU-60, it achieves better transfer performance for action localization on PKUMMD.  


\begin{figure*}[t]
\centering
\includegraphics[width=0.9\textwidth]{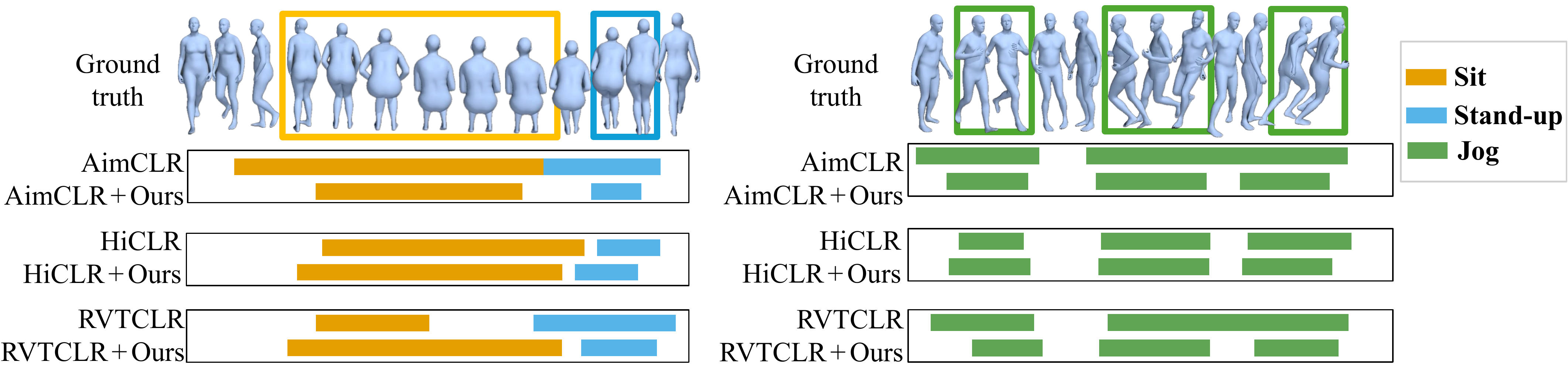}
\caption{{\bf Qualititive visualization of action predictions on BABEL.} 
We compare the ground truth with predictions from three baselines with and without  our approach. 
(Left) there is a brief transition between \textit{Sit} and \textit{Stand-up}, where the subject leans forward in preparation. (Right) the subject performs the \textit{Jog} action multiple times in different directions, with turns separating each action instance. }
\label{fig:bars}
\end{figure*}

\noindent{\bf Qualititive Evaluation --} We visualize examples of detected actions in Fig.~\ref{fig:bars}, comparing predictions from baselines with and without our method. For the actions \textit{Sit} and \textit{Stand-up}, combining with our approach results in more precise boundaries for all baselines. For example, RVTCLR predicts the action \textit{Sit} mainly at the onset, capturing only the most distinctive moment when the subject begins {sitting}. While this is sufficient for video-level recognition, it fails to model the full temporal extent. Combined with our method, the prediction covers the rest of the {sitting} phase, providing a more accurate boundary at the end of the action.  For \textit{Stand-up}, AimCLR and RVTCLR produce inaccurate start boundaries, while our method helps identify the sit-to-stand onset more precisely. The \textit{Jog} example contains several jogging actions in different directions, separated by turning in place as transitions. AimCLR and RVTCLR fail to distinguish the transition phase and instead classify it as part of the \textit{Jog} action, whereas when augmented with ours, the short transition is correctly excluded. 
{This can be explained by the snippet contrastive pretraining, which promotes fine-grained discrimination to enhance temporal sensitivity}\footnote{\scriptsize {We provide visualizations of the pretrained feature distributions in the Suppl. Materials.}}.


\begin{table}[t]
\caption{\textbf{Ablations on the proposed designs and the hyperparameter.} mAP over tIoU=0.1 are reported under KNN evaluation. {Here, 'w/o both' denotes  baseline method RVTCLR~\cite{RVTCLR}, while 'dense projection' and 'feature fusion' refers to our designs described in \ref{sec:dense contrast} and \ref{sec: unet}.}} 
\centering
\resizebox{\textwidth}{!}{%
\begin{tabular}{c|cccc|cccc|cccc|c}
\toprule
\multirow{2}{*}{\textbf{Settings}} & \multicolumn{4}{c|}{\textbf{Subset-1}} & \multicolumn{4}{c|}{\textbf{Subset-2}} & \multicolumn{4}{c|}{\textbf{Subset-3}}& \\
&\textit{Walk} & \textit{Stand} & \textit{Turn} & \textit{Jump} & \textit{Sit} & \textit{Run} & \textit{Stand-up} & \textit{Kick} & \textit{Jog} & \textit{Wave} & \textit{Dance} & \textit{Gesture} &\cellcolor{gray!10}\textbf{Avg.}\\
\midrule

w/o both                    
& 71.1 & 69.3 & 21.8 & \underline{30.5} 
& 68.9 & 43.2 & 36.2 & \underline{61.1} 
& 53.2 & 19.9 & 31.9 & 11.7 
& \cellcolor{gray!10}43.2 \\

& \underline{72.5} & \underline{68.1} & \underline{26.9} & 28.8 
& 72.4 & 35.3 & \underline{49.2} & 54.1 
& 57.7 & \underline{26.1} & 27.8 & \underline{18.9} 
& \cellcolor{gray!10}\underline{44.8} \\

w/ feature fusion        
& 71.5 & 58.5 & 15.3 & 28.1 
& 74.1 & \textbf{53.2} & 34.1 & 35.3 
& 58.4 & \textbf{28.3} & \underline{32.8} & \textbf{28.1} 
& \cellcolor{gray!10}43.4 \\

\midrule

Ours       
& \textbf{75.3} & \textbf{69.4} & \textbf{31.3} & \textbf{41.9} 
& \textbf{73.4} & \underline{46.4} & \textbf{53.1} & \textbf{64.0} 
& \textbf{68.1} & 18.8 & \textbf{34.8} & 14.0 
& \cellcolor{gray!10}\textbf{49.2} \\

\bottomrule
\end{tabular}
}
\label{tab:ablation_design}
\vspace{-20pt}
\end{table}

\section{Ablations}
\noindent{\bf Proposed Design --} As shown in Tab.~\ref{tab:ablation_design}, removing either the dense snippet contrastive objective (Section~\ref{sec:dense contrast}), the multiscale feature fusion module (Section~\ref{sec: unet}), or both, leads to clear performance drops, reducing the average mAP from 49.2\% to 43.2\% ($\downarrow$6.0\%).  {The extra temporal sensitivity introduced by dense projection improves localization for \textit{Jog}, which exhibits relatively stable temporal patterns. In contrast, \textit{Run} shows much higher intra-class variability in pace and motion dynamics, making dense temporal contrast less effective. Nevertheless, the broader context captured by our U-shaped feature fusion benefits both actions.  For \textit{Wave} and \textit{Gesture}, while each proposed design improves localization performance on these actions when applied separately, their combination leads to a drop.} 
Unlike whole-body actions, e.g. \textit{Walk} or \textit{Dance}, where pronounced details help refine boundaries for localization, \textit{Wave} and \textit{Gesture} rely on subtle hand movements that are easily {missed}  when local details are overemphasized. 





\begin{table}[t]
\centering
\scriptsize

\begin{minipage}[t]{0.4\textwidth}
\centering
\vspace{0pt}
\caption{\textbf{Ablation on $\lambda$.} Averaged mAP across tIoU from 0.1 to 0.5 under KNN evaluation.}
\begin{tabular}{c|ccc|c}
\toprule
\multirow{2}{*}{\textbf{$\lambda$}} & \multicolumn{3}{c|}{\textbf{mAP@tIoU (\%)}} & \\
 & Subset-1 & Subset-2 & Subset-3 & \cellcolor{gray!10}\textbf{Avg.} \\
\midrule
0.5  & 43.0 & 44.1 & 28.3 & \cellcolor{gray!10}{38.5} \\
1    & \underline{43.3} & \textbf{46.9} & \underline{30.2} & \cellcolor{gray!10}{\underline{40.1}} \\
1.5  & \textbf{43.4} & \underline{46.3} & \textbf{31.2} & \cellcolor{gray!10}{\textbf{40.3}} \\
2    & 42.7 & 42.2 & 30.0 & \cellcolor{gray!10}{38.3} \\
\bottomrule
\end{tabular}
\label{tab:ablation_lambda}
\end{minipage}
\hfill
\begin{minipage}[t]{0.5\textwidth}
\centering
\vspace{0pt}
\caption{\textbf{Ablation on $N$ and snippet length.} Averaged mAP across tIoU from 0.1 to 0.5 under KNN evaluation.}
\begin{tabular}{cc|ccc|c}
\toprule
\multirow{2}{*}{\textbf{N}} & \multirow{2}{*}{\textbf{Length}} &
\multicolumn{3}{c|}{\textbf{mAP@tIoU (\%)}} & \\
 & & Subset-1 & Subset-2 & Subset-3 & \cellcolor{gray!10}\textbf{Avg.} \\
\specialrule{0.15em}{0pt}{0pt}
 & - & 40.2 & 40.9 & 27.5 & \cellcolor{gray!10}{36.2} \\
1 & 300 & 44.5 & 43.3 & 29.1 & \cellcolor{gray!10}{39.0} \\
3 & 100 & 43.2 & 43.1 & 27.2 & \cellcolor{gray!10}{37.9} \\
5 & 64 & 41.1 & 41.0 & \underline{30.0} & \cellcolor{gray!10}{37.4} \\
10 & 32 & \textbf{45.6} & 45.1 & 29.3 & \cellcolor{gray!10}{\underline{40.0}} \\
19 & 16 & 43.4 & \textbf{46.3} & \textbf{31.2} & \cellcolor{gray!10}{\textbf{40.3}} \\
38 & 8 & \underline{44.5} & 43.9 & 29.7 & \cellcolor{gray!10}{39.4} \\
\bottomrule
\end{tabular}
\label{tab:ablation_N}
\end{minipage}
\vspace{-20pt}
\end{table}

\noindent{\bf Loss weight $\lambda$ --}  
Overemphasizing the snippet-level loss degrades performance, since the randomly initialized model at the start of pretraining cannot yet generate reliable correspondences for cross-view positive snippet matching. 
Setting $\lambda=1.5$ (see Tab.~\ref{tab:ablation_lambda}) achieves the best trade-off, with average mAP$=40.3\%$.

\noindent{\bf Snippet embeddings $N$ --}  
For a fair comparison, we follow~\cite{aimclr22,RVTCLR} and fix the input to 300 frames.  As shown in Tab.~\ref{tab:ablation_N}, both too many and too few snippets are suboptimal, with 19 snippets of 16 frames achieving the best overall average of 40.3\%. 
The optimal $N$ for each subset is different due to variations in action lengths and characteristics. For example, on Subset-3, using 5 snippets of 64 frames yields 30.0\% mAP, which is higher than the 29.3\% obtained with 10 snippets of 32 frames, while on the other subsets the result is lower. When $N=1$, our method is equivalent to the video-level baseline, except that the dense projection module applies a nonlinear activation function, with a separate memory bank for snippet embeddings. 
\vspace{-5pt}

\section{Conclusion}
\label{sec: conclusion}
\vspace{-2pt}
We introduced a snippet-level discrimination pretext task for skeleton-based self-supervised contrastive learning to preserve temporal sensitivity for downstream action localization. Experiments on BABEL and PKUMMD
demonstrate the snippet-level contrast improves the performance of existing skeleton-based contrastive learning methods. 
We also introduced a U-shaped plug-and-play module that enables dense localization for existing skeleton-based action recognition backbones. Our method reduces reliance on dense frame-level labels, achieving performance comparable to training from scratch in weakly supervised setting and surpassing it in the semi-supervised setting.
Finally, our transfer learning results show that pretraining on long and untrimmed skeleton sequences{, such as BABEL,} 
is more effective for localization than pretraining on curated, short skeleton sequences{, such as NTU-60.}
{Future work could explore the transfer learning capacity of snippet contrastive pretraining  for other motion-related tasks, such as action quality assessment~\cite{majid24pecop} or motion generation~\cite{majid25gaitgen}, as well as leveraging skeleton data that are more domain-specific \qsd{and label-demanding}
such as Parkinson's Disease gait sequences~\cite{majid2023remap,majid25yourturn}.}
\subsubsection{Acknowledgements}
This work was supported by the TORUS Project, which has been funded by the UK Engineering and Physical Sciences Research Council (EPSRC), grant number EP/X036146/1. The first author is supported by EPSRC (EP/S023704/1).

%
%
%

\bibliographystyle{splncs04}
\bibliography{references}
%




\title{Supplementary materials for "Skeleton-Snippet Contrastive Learning with  Multiscale Feature Fusion for Action Localization"}

\author{Q. Cheng\inst{1}\and
J. Liu\inst{1}\and
C. Morgan\inst{2} \and 
A. Whone\inst{2}\and
M. Mirmehdi\inst{1}}

\authorrunning{Q. Cheng et al.}
%
\institute{School of Computer Science, University of Bristol\and North Bristol NHS Trust}
\maketitle     
\renewcommand{\thetable}{S\arabic{table}}

\section{KNN and Linear Evaluation with Different Backbones}

In the main paper's Tab. 3 and Tab. 4, we evaluate our approach using different CL baseline methods. Here, we fix the baseline method to RVTCLR~\cite{RVTCLR} and investigate the effect of varying the backbone architecture, including ST-GCN~\cite{stgcn18}, 2s-AGCN~\cite{2sagcn19}, and UNIK~\cite{unik21}. As shown in Tab.~\ref{tab:knn_with_different_backbones}, our method consistently improves the performance of 2s-AGCN and UNIK across all subsets under the KNN evaluation protocol. For ST-GCN, however, we observe a slight performance drop on Subset-2 and Subset-3. We believe this is because ST-GCN backbone is a basic model with lower learning capacity compared to the other models, which limits the benefit of our complex training strategies when using direct KNN clustering in the pretrained feature space for frame-level localization. Under linear evaluation (see Tab.~\ref{tab: linear_different_backbone}), as the linear classification head has greater modelling capacity than KNN, the performance improves consistently across all backbones and subsets.

\begin{table}[h]
\caption{{\bf KNN evaluation with different backbones.} mAP averaged over actions at tIoU thresholds 0.1 to 0.5 on three subsets of BABEL.}
\centering
\renewcommand{\arraystretch}{1.3}
\setlength{\tabcolsep}{2pt}
\resizebox{\linewidth}{!}{
\begin{tabular}{c|c|ccccc>{\columncolor{gray!10}}c|ccccc>{\columncolor{gray!10}}c|ccccc>{\columncolor{gray!10}}c}
\toprule
\multirow{2}{*}{\textbf{Method}}  &\multirow{2}{*}{\textbf{Backbone}}  
& \multicolumn{6}{c|}{\textbf{Subset-1}} & \multicolumn{6}{c|}{\textbf{Subset-2}} & \multicolumn{6}{c}{\textbf{Subset-3}}\\
&& 0.1& 0.2& 0.3& 0.4& 0.5 & \textbf{Avg.} & 0.1& 0.2& 0.3& 0.4& 0.5& \textbf{Avg.} & 0.1& 0.2& 0.3& 0.4& 0.5 &\textbf{Avg.}\\
\midrule
RVTCLR~\cite{RVTCLR}&\multirow{2}{*}{ST-GCN~\cite{stgcn18}}&43.3&39.3&34.9&29.9&26.2&34.7&\textbf{46.4}&\textbf{41.2}&\textbf{32.1}&\textbf{25.1}&17.0&\textbf{32.3}&\textbf{32.0}&\textbf{22.8}&\textbf{16.8}&14.3&12.1&\textbf{19.6}\\
+Ours& &\textbf{46.3}&\textbf{42.0}&\textbf{36.4}&\textbf{32.3}&\textbf{28.3}&\textbf{37.1}&44.1&37.3&30.7&23.0&\textbf{19.1}&30.9&25.3&19.9&16.7&\textbf{15.5}&\textbf{13.8}&18.3\\
\midrule
RVTCLR~\cite{RVTCLR}&\multirow{2}{*}{2s-AGCN~\cite{2sagcn19}} & 48.2 & 43.5 & 39.6 & 33.1 & 26.8 & 38.2 & 52.3 & 44.2 & 35.7 & 27.4 & 21.8 & 36.3 & 29.2 & 24.4 & 21.6 & 20.2 & 18.0 & 22.7\\
+Ourss& &\textbf{52.6} & \textbf{48.1} & \textbf{44.5} & \textbf{39.2} & \textbf{32.2} & \textbf{43.4} & \textbf{59.6} & \textbf{55.0} & \textbf{46.2} & \textbf{40.3} & \textbf{30.1} & \textbf{46.3} & \textbf{42.0} & \textbf{34.3} & \textbf{29.8} & \textbf{25.6} & \textbf{24.2} & \textbf{31.2}\\
\midrule
RVTCLR~\cite{RVTCLR}&\multirow{2}{*}{UNIK~\cite{unik21}}&47.3&42.8&37.8&\textbf{33.2}&\textbf{28.7}&\textbf{38.0}&\textbf{49.2}&\textbf{43.2}&35.2&26.9&23.2&35.6&\textbf{31.3}&21.8&19.6&16.0&\textbf{14.9}&20.7\\
+Ours& &\textbf{48.3}&\textbf{43.2}&\textbf{38.0}&\textbf{32.9}&27.3&37.9 &48.4&42.6&\textbf{36.8}&\textbf{29.0}&\textbf{23.3}&\textbf{36.0} &31.1&\textbf{25.2}&\textbf{20.9}&\textbf{18.0}&14.3&\textbf{21.9}\\
\bottomrule
\end{tabular}}
\label{tab:knn_with_different_backbones}
\end{table}

\begin{table}[h]
\caption{{\bf Linear evaluation with different backbones.} mAP averaged over actions at tIoU thresholds 0.1 to 0.5 on three subsets of BABEL.}
\centering
\renewcommand{\arraystretch}{1.3}
\setlength{\tabcolsep}{2pt}
\resizebox{\linewidth}{!}{
\begin{tabular}{c|c|ccccc>{\columncolor{gray!10}}c|ccccc>{\columncolor{gray!10}}c|ccccc>{\columncolor{gray!10}}c}
\toprule
\multirow{2}{*}{\textbf{Method}}  &\multirow{2}{*}{\textbf{Backbone}}  
& \multicolumn{6}{c|}{\textbf{Subset-1}} & \multicolumn{6}{c|}{\textbf{Subset-2}} & \multicolumn{6}{c}{\textbf{Subset-3}}\\
&& 0.1& 0.2& 0.3& 0.4& 0.5 & \textbf{Avg.} & 0.1& 0.2& 0.3& 0.4& 0.5& \textbf{Avg.} & 0.1& 0.2& 0.3& 0.4& 0.5 &\textbf{Avg.}\\
\midrule
RVTCLR~\cite{RVTCLR}&\multirow{2}{*}{ST-GCN~\cite{stgcn18}} &59.0&54.4&47.5&38.8&30.4&46.0&63.4&52.1&42.1&34.1&24.4&43.2&46.7&43.7&38.9&31.5&26.1&37.4\\
+Ours& &\textbf{60.0}&\textbf{55.9}&\textbf{49.7}&\textbf{39.8}&\textbf{32.0}&\textbf{47.5}&\textbf{65.2}&\textbf{58.1}&\textbf{50.5}&\textbf{42.0}&\textbf{32.1}&\textbf{49.6}&\textbf{52.1}&\textbf{50.4}&\textbf{45.0}&\textbf{37.3}&\textbf{34.5}&\textbf{43.9}\\
\midrule
RVTCLR~\cite{RVTCLR}&\multirow{2}{*}{2s-AGCN~\cite{2sagcn19}} &57.0&50.1&41.9&31.8&23.7&40.9&57.7&48.5&38.2&27.9&20.7&38.8&43.3&40.4&34.2&29.8&26.5&34.8\\ 
+Ours& &\textbf{60.5}&\textbf{55.0}&\textbf{45.8}&\textbf{36.6}&\textbf{28.8}&\textbf{45.3}&\textbf{65.2}&\textbf{57.8}&\textbf{49.0}&\textbf{40.4}&\textbf{32.9}&\textbf{49.1}&\textbf{47.9}&\textbf{46.8}&\textbf{41.7}&\textbf{31.7}&\textbf{28.2}&\textbf{39.3}\\ 
\midrule
RVTCLR~\cite{RVTCLR}&\multirow{2}{*}{UNIK~\cite{unik21}}&\textbf{61.9}&\textbf{55.3}&47.1&39.3&30.6&46.9&62.8&57.3&48.6&39.9&31.4&48.0&\textbf{47.6}&43.7&38.3&33.3&\textbf{31.5}&\textbf{38.9}\\

+Ours& &59.7&\textbf{55.3}&\textbf{49.0}&\textbf{39.8}&\textbf{31.7}&\textbf{47.1}&\textbf{67.4}&\textbf{59.3}&\textbf{51.1}&\textbf{42.5}&\textbf{32.5}&\textbf{50.6}&46.0&\textbf{43.8}&\textbf{39.1}&\textbf{33.7}&30.7&38.7\\
\bottomrule
\end{tabular}}
\label{tab: linear_different_backbone}
\end{table}

\section{Additional Qualitative Evaluation}
We also visualize the pretrained feature distributions in Fig.~\ref{fig:sidebyside} 
at fixed hyperparameter settings for RVTCLR~\cite{RVTCLR} and RVTCLR+ours.
\renewcommand{\thefigure}{S\arabic{figure}}
\begin{figure}[h]
\centering
\includegraphics[width=\textwidth]{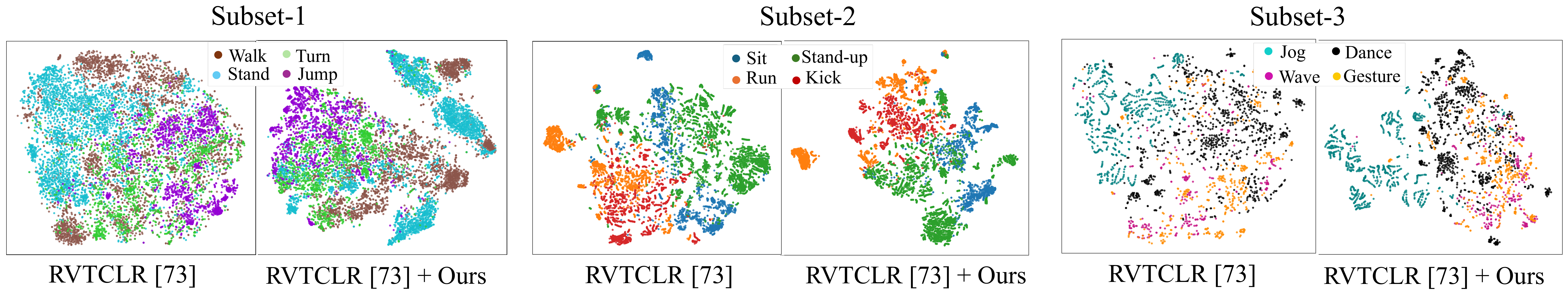}
\caption{{{t-SNE visualization of pretrained features on three BABEL subsets.} 
{Each point represents the pretrained feature of a single frame after downsampling, which shows how frames from similar actions across videos group together in the feature space.}}}
\label{fig:sidebyside}
\end{figure}

The plots show that incorporating our method leads to tighter clusters and improved feature discriminability, e.g.  for \textit{Stand} in Subset-1,  \textit{Sit} in Subset-2, and  \textit{Jog} in Subset-3. For example, while \textit{Stand} still appears in multiple clusters after applying our method, these clusters are more compact and distinct compared to those obtained using RVTCLR~\cite{RVTCLR} alone. This suggests that our snippet-level fine-grained contrastive pretraining leads to multiple sub-groups (e.g., different postures or contexts of standing) and improves the overall separability from other classes. {The figure also shows that the subtle local motions, \textit{Wave} and \textit{Gesture} remain relatively entangled compare to full-body movements such as \textit{Jog} or \textit{Jump}. This finding is consistent with the ablation results in the main paper. }








\end{document}